\documentclass{article}

\usepackage[preprint]{neurips}
\usepackage[utf8]{inputenc} 
\usepackage[T1]{fontenc}    
\usepackage{hyperref}       
\usepackage{url}            
\usepackage{booktabs}       
\usepackage{amsfonts}       
\usepackage{nicefrac}       
\usepackage{microtype}      
\usepackage{pifont}

\usepackage{neurips}
\usepackage{times}
\usepackage{latexsym}
\usepackage{amsmath}
\usepackage{amsfonts}
\usepackage[linesnumbered,ruled,vlined]{algorithm2e}
\usepackage{graphicx} 
\usepackage{booktabs}
\usepackage{array}
\usepackage{enumitem}
\usepackage{amsthm}
\usepackage{algpseudocode}
\usepackage[switch]{lineno}
\usepackage[utf8]{inputenc}
\usepackage{eqparbox}
\usepackage[nopar]{lipsum}
\usepackage{multirow}
\usepackage{makecell}
\usepackage{xcolor}
\usepackage{hhline} 
\usepackage{microtype}

\usepackage{rotating}
\usepackage{geometry}
\usepackage{pdflscape}

\usepackage{adjustbox}
\usepackage{array}
\usepackage{booktabs}
\usepackage{multirow}
\usepackage{colortbl}

\usepackage{relsize}

\usepackage{booktabs,multirow,array}
\newcolumntype{N}{@{}m{0pt}@{}}

\newcolumntype{R}[2]{%
    >{\adjustbox{angle=#1,lap=\width-(#2)}\bgroup}%
    l%
    <{\egroup}%
}
\newcommand*\rot{\multicolumn{1}{R{45}{1em}}}

\usepackage[many]{tcolorbox}
\newtcolorbox{fancyquotes}{%
    enhanced jigsaw, 
    breakable,      
    frame hidden,   
    left=0.5cm,       
    right=0.1cm,      
    overlay={%
        \node [scale=8,
            text=black,
            inner sep=0pt,] at ([xshift=-1cm,yshift=-1cm]frame.north west){}; 
        \node [scale=8,
            text=black,
            inner sep=0pt,] at ([xshift=1cm]frame.south east){};  
            },
                parbox=false,
}

\usepackage{mathtools, nccmath}
\usepackage{scrextend}
\deffootnote[.25in]{.25in}{.15in}{\makebox[.25in][r]{\thefootnotemark .\hspace{.15in}}}

\makeatletter

\newtheorem*{proof*}{Proof}

\definecolor{Gray}{gray}{0.85}
\definecolor{LightCyan}{rgb}{0.88,1,1}

\newcolumntype{a}{>{\columncolor{Gray}}c}
\newcolumntype{b}{>{\columncolor{white}}c}

\usepackage{listings}
\usepackage{color}
\definecolor{codegreen}{rgb}{0.3,0.5,0.0}
\def\@fnsymbol#1{\ensuremath{\ifcase#1\or \dagger\or *\or \ddagger\or
   \mathsection\or \mathparagraph\or \|\or **\or \dagger\dagger
   \or \ddagger\ddagger \else\@ctrerr\fi}}

\newcolumntype{C}[1]{>{\centering\let\newline\\\arraybackslash\hspace{0pt}}m{#1}}
\newcommand\ChangeRT[1]{\noalign{\hrule height #1}}

\title{Masked Autoencoders As The Unified Learners For Pre-Trained Sentence Representation}

\author{Alexander Liu$^\dag$, Samuel Yang$^\ddag$ \\
  $\dag$: The Hong Kong University of Science and Technology, Kowloon, Hong Kong \\
  $\ddag$: University of Science and Technology of China, Hefei, China \\
  \texttt{$\dag$:axliu.hkust@outlook.com,
  $\ddag$:samuelyang150@gmail.com}
} 

\begin{document}
\maketitle 

\begin{abstract}

Despite the progresses on pre-trained language models, there is a lack of unified frameworks for pre-trained sentence representation. As such, it calls for different pre-training methods for specific scenarios, and the pre-trained models are likely to be limited by their universality and representation quality. In this work, we extend the recently proposed MAE style pre-training strategy, RetroMAE \citep{liu2022retromae}, such that it may effectively support a wide variety of sentence representation tasks. 
The extended framework consists of two stages, with RetroMAE conducted throughout the process. The first stage performs RetroMAE over generic corpora, like Wikipedia, BookCorpus, etc., from which the {base model} is learned. The second stage takes place on domain-specific data, e.g., MS MARCO and NLI, where the base model is continuingly trained based on RetroMAE and contrastive learning. The pre-training outputs at the two stages may serve different applications, whose effectiveness are verified with comprehensive experiments. Concretely, the base model are proved to be effective for zero-shot retrieval, with remarkable performances achieved on BEIR benchmark \citep{thakur2021beir}. The continuingly pre-trained models further benefit more {downstream tasks}, including the domain-specific dense retrieval on {MS MARCO} \citep{nguyen2016ms}, {Natural Questions} \citep{kwiatkowski2019natural}, and the sentence embeddings' quality for standard STS \citep{agirre2016semeval,cer2017semeval,marelli2014sick} and transfer tasks in SentEval \citep{conneau2018senteval}. The empirical insights of this work may inspire the future design of sentence representation pre-training. Our pre-trained models and source code will be released to the public communities.


\end{abstract} 


\section{Introduction}
Sentence representation is a fundamental issue in natural language processing. It transforms each input sentence into its embedding, where semantically similar instances can be located close to each other in the latent space. Thanks to this property, sentence representation models are utilized in many important applications, such as semantic matching, dense retrieval, and recommender systems. 
With the development of deep learning, deep neural networks are widely utilized for sentence representation tasks. However, the training of deep models usually call for massive scales of labeled data, which can be prohibitive in many real-world scenarios. 

The pre-trained language models are regarded as a promising way to alleviate the data scarcity problem. In recent years, the pre-trained language models are specifically tailored for sentence representation purposes. For example, there have been pre-training methods developed for dense retrievers \citep{chang2020pre,guu2020realm,gao2021condenser,gao2021unsupervised,lu2021less}; besides, there have also been models pre-trained for tasks like semantic matching \citep{reimers2019sentence,thakur-2020-AugSBERT,gao2021simcse}. However, the existing pre-trained models rely on differentiated pre-training workflow, where a pre-trained model for one specific task contributes little to other scenarios. Such a problem is unfavorable to both universality and representation quality; indeed, a unified pre-training framework will be highly desirable for sentence representation. 

The MAE-style algorithms become increasingly popular recently: it was recognized to be a strong pre-trainer for various tasks in computer vision communities \citep{he2022masked,feichtenhofer2022masked}; later on, it was also proved to be extremely effective for zero-shot dense retrieval in RetroMAE \citep{liu2022retromae}. In this work, we demonstrate that RetroMAE may serve as the unified pre-training framework for sentence representation, as it can be easily extended to support various scenarios. The extension is inspired by the previous practice on generic pre-training algorithms \citep{gururangan2020don}, where a two-stage workflow is conducted. The first stage training takes place on generic corpora, like Wikipedia and BookCorpus, where the base model is learned through RetroMAE. The second stage leverages domain-specific corpora, such as passages from MS MARCO \citep{nguyen2016ms} and sentence pairs from NLI datasets \citep{bowman2015large,williams2017broad}, where the model is continuingly trained through RetroMAE and contrastive learning. The extended pre-training workflow is conceptually simple and empirically competitive, where the generated models may achieve remarkable performances in different application scenarios. Particularly, the following empirical insights are derived from our experiments.  

\begin{itemize}[leftmargin=5.0mm]
    \item The \textbf{RetroMAE pre-training on generic corpora} is critical, as it initializes the model with a strong capacity on semantic encoding. Our reproduction of the original RetroMAE  remarkably achieves \textbf{0.452} on \textbf{BEIR benchmark} in total average, which is way beyond any existing models with similar model sizes and pre-training data.

    \item The \textbf{continued in-domain pre-training} (also known as {post pre-training}) is important to promote downstream tasks' performances. For example, the retrieval quality (with simple ANCE style fine-tuning) can be notably lifted to \textbf{0.393} (MRR@10) and \textbf{0.985} (Recall@1000) for \textbf{MS MARCO} passage retrieval task, compared with 0.378 (MRR@10) and 0.980 (Recall@1000) from the base model pre-trained on generic corpora. These results are highly competitive on the corresponding benchmark, which are even comparable to those based on much more expensive knowledge distillation \citep{ren2021rocketqav2} and adversarial learning \citep{zhang2021adversarial}. 
    \item The \textbf{contrastive learning}'s impact is conditional. On one hand, it helps to generate high quality sentence embeddings for standard \textbf{STS} \citep{agirre2016semeval,cer2017semeval,marelli2014sick} and \textbf{transfer tasks} in {SentEval} \citep{conneau2018senteval}; on the other hand, it is found to bring limited benefit to dense retrieval tasks. We conjecture that it's the quality of relevance label that makes the difference. Traditional self-contrastive learning \citep{chang2020pre,guu2020realm}, which rely on data augmentation to generate positive samples, are prone to inferior label quality. Thus, they could be in vain at the presence of stronger pre-trainers, like RetroMAE. 
\end{itemize} 

To conclude, the main points of this paper are summarized as follows. We extend RetroMAE with continued in-domain pre-training and contrastive learning, such that it may contribute to various sentence representation tasks, including zero-shot retrieval, domain-specific retrieval, standard STS and transfer tasks in SentEval. We perform comprehensive experimental studies, whose results reveal the impacts introduced by RetroMAE, continued in-domain pre-training, and contrastive learning. The empirical insights may inspire the design of sentence representation pre-training in the future. 

%


\section{Methodology}
In this section, we'll present the preliminaries about RetroMAE in the first place. Then, we'll make discussions about our effort to extend it for different downstream tasks. 

\subsection{Preliminaries}
\begin{figure*}[t]
\centering
\includegraphics[width=1.0\textwidth]{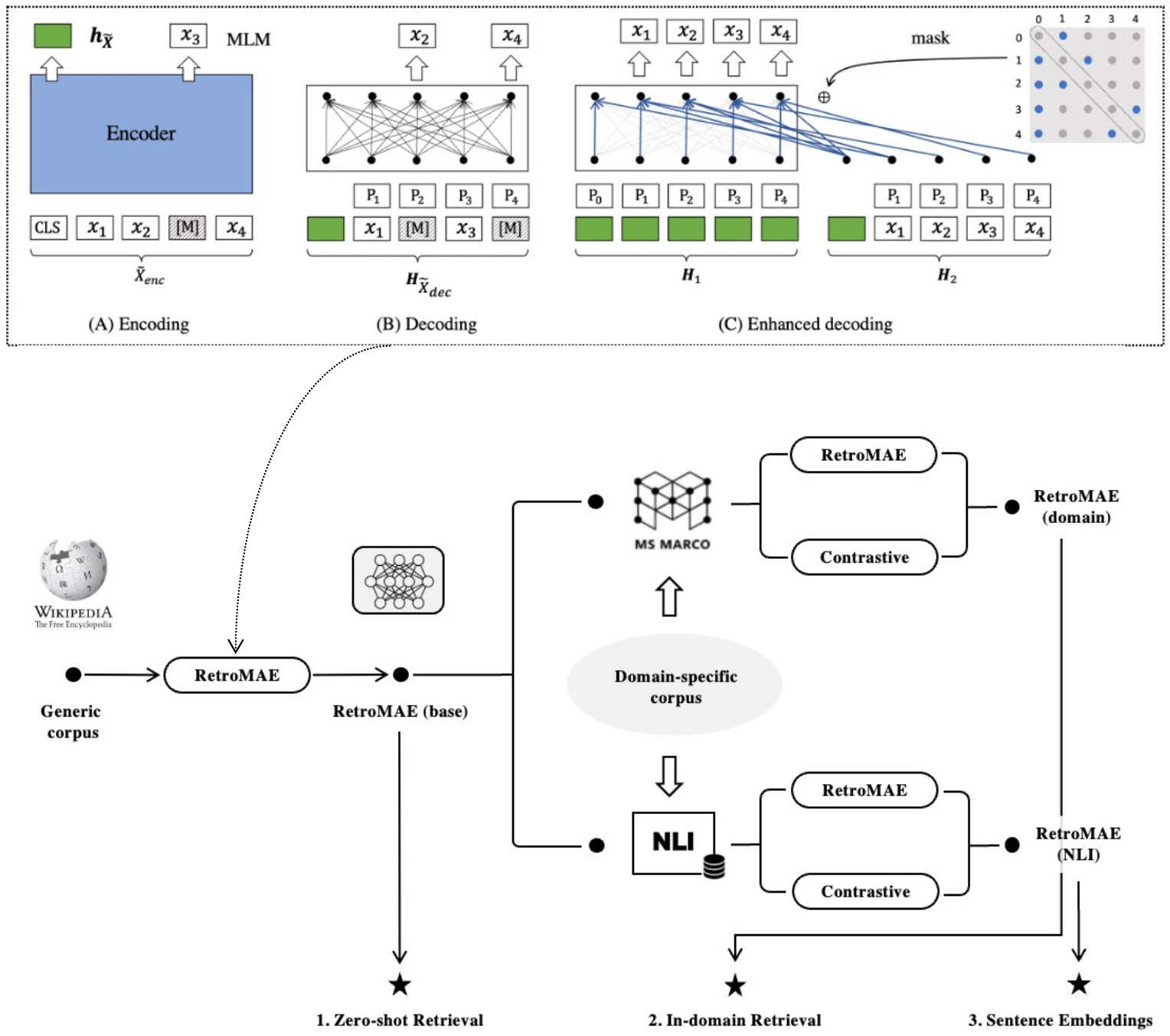}
\caption{The two-stage pre-training workflow with RetroMAE. The first stage performs RetroMAE on generic data; the base model will effectively support tasks like zero-shot retrieval. The second stage takes place on domain-specific data; the base model is further trained on domain-specific data with RetroMAE and contrastive learning, which benefits tasks like in-domain retrieval and STS.} 
\label{fig:1}
\end{figure*}

RetroMAE is to encode the masked sentence into its embedding, and then recover the original sentence based on the masked input and the sentence embedding. It is highlighted for two features: 1) the asymmetric model structures, and 2) the asymmetric masking ratios about the encoder and decoder. The training workflow of RetroMAE is shown as Figure \ref{fig:1}. For the \textbf{encoding} operation, the input sentence $X$ is masked as $\tilde{X}_{enc}$, with a small fraction of its tokens randomly replaced by the special token [M] (the shadowed rectangles). The masked input is transformed into sentence embedding $\mathbf{h}_{\tilde{X}}$ based on the encoding network $\Phi^{enc}(\cdot)$: 
\begin{equation}
    \mathbf{h}_{\tilde{X}} \leftarrow \Phi_{enc}(\tilde{X}_{enc}). 
\end{equation}
Note that a relatively moderate masking ratio is applied to the encoding operation, where 15$\sim$30\% of the input tokens are sampled for masking. The encoder is a BERT-base scale transformer so as to ensure the quality of sentence representation.

The \textbf{decoding} step is to recover the original sentence from the sentence embedding and the masked input. Particularly, the input sentence $X$ is polluted again for decoder as $\tilde{X}_{dec}$ with a different mask. The masking ratio is chosen to be more aggressive than the one used by encoder, with 50$\sim$70\% of the input tokens replaced by the special token [M]. The sentence embedding from the encoder and the masked input of the decoder are combined into the following sequence:  
\begin{equation}
    \mathbf{H}_{\tilde{X}_{dec}} \leftarrow [\mathbf{h}_{\tilde{X}}, \mathbf{e}_{x_1}+\mathbf{p}_1, ... , \mathbf{e}_{x_N}+\mathbf{p}_N],
\end{equation}
where $\mathbf{e}_{x_i}$ stands for the embedding of $x_i$, to which an extra position embedding $\mathbf{p}_i$ is added.
The decoder $\Phi_{dec}$ is learned to recover the original sentence $X$ by minimizing the following loss function: 
\begin{equation}
    \mathcal{L}_{dec} = \sum_{x_i \in \text{masked}} \textit{cross-entropy}(x_i|\Phi_{dec}( \mathbf{H}_{\tilde{X}_{dec}}) ).
\end{equation}
One notable feature about the decoding network is that it merely consists of a one-layer transformer. For one thing, it may save a great deal of computation cost; for another thing, it will ensure the decoding operation to be challenging enough, which forces the generation of high-quality sentence embedding from the encoder.

The basic decoding may only derive training signals (the cross-entropy losses) from the masked tokens; besides, every masked token is always reconstructed based on the same context, i.e., $\mathbf{H}_{\tilde{X}_{dec}}$. As a result, it is prone to low sample efficiency. To mitigate this problem, the \textbf{enhanced decoding} is introduced, where training signals can be derived from the entire input sentence, and the prediction of the masked tokens can be performed based on {diversified contexts}. Particularly, there are two input sequences: $\mathbf{H}_1$ (query stream) and $\mathbf{H}_2$ (context stream), generated for the decoding operation: 
\begin{equation}
\begin{gathered}
\label{eq:4}
\mathbf{H}_1 \leftarrow [\mathbf{h}_{\tilde{X}} + \mathbf{p}_0,...,
\mathbf{h}_{\tilde{X}} + \mathbf{p}_N], \\
\mathbf{H}_2 \leftarrow  
[\mathbf{h}_{\tilde{X}}, \mathbf{e}_{x_1}+\mathbf{p}_1, ... , \mathbf{e}_{x_N}+\mathbf{p}_N]. 
\end{gathered}
\end{equation}
in which $\mathbf{h}_{\tilde{X}}$ is the sentence embedding from encoder, $\mathbf{e}_{x_i}$ is the token embedding (all tokens are left non-masked in the place), and $\mathbf{p}_i$ is the position embedding.
Let $\mathbf{M} \in \mathbb{R}^{L \times L}$ be the position-specific attention mask, with which the self-attention operation is conducted as follows:
\begin{equation}\label{eq:5}
\begin{gathered}
    \mathbf{Q} = \mathbf{H}_1\mathbf{W}^Q, \mathbf{K} = \mathbf{H}_2\mathbf{W}^K, \mathbf{V} = \mathbf{H}_2\mathbf{W}^V; \\
    \mathbf{M}_{ij} = 
    \begin{cases}
    0, ~~~~~~\text{can be attended}, \\
    -\infty, ~\text{masked}; 
    \end{cases} \\
    \mathbf{H}^o_{dec} = \mathrm{softmax}(\frac{\mathbf{Q}^T\mathbf{K}}{\sqrt{d}} + \mathbf{M
    })\mathbf{V}. 
\end{gathered}
\end{equation}
The decoding output $\mathbf{H}^o_{dec}$ is used to reconstruct the original sentence (operations like layer-norm, FFN, and residual connections are omitted for simplicity). Finally, the decoder is learned to optimize the following objective: 
\begin{equation}\label{eq:6}
 \mathcal{L}_{dec} = \sum\nolimits_{x_i \in X} \textit{cross-entropy}(x_i|\mathbf{H}^o_{dec}).
\end{equation} 
Because the decoding network has one single layer, each token $x_i$ can only be recovered from the context visible to the $i$-th row of the matrix $\mathbf{M}$, which are generated by the following rules: 
\begin{equation}\label{eq:7}
\begin{gathered}
\mathbf{M}_{ij} = 
    \begin{cases}
    0,   ~~ x_j \in s(X_{\neq i}), ~\text{or}~ j_{{|i\neq0}}=0 \\
    -\infty,  ~~ \text{otherwise}. 
    \end{cases} 
\end{gathered}
\end{equation}
The sampled tokens, $s(X_{\neq i})$, and the first elements (except the first row) will be visible when recovering $x_i$. All diagonal elements, i.e., $x_i$ in the $i$-th row, will be excluded, such that each token may not attend to itself during decoding. 

As mentioned, the first stage pre-training performs RetroMAE on generic data, like Wikipedia and BookCorpus. It worth noting that the masked tokens for the encoding network will also be reconstructed in the same way as the conventional masked language modeling (MLM). The corresponding loss is denoted as $\mathcal{L}_{enc}$. The encoder and decoder are jointly trained to minimize the summation of $\mathcal{L}_{dec}$ and $\mathcal{L}_{enc}$. Finally, the well-trained encoding network is utilized as the base model, which will serve applications like zero-shot dense retrieval.

\begin{table*}[t]
    \centering
    \tiny
    \begin{tabular}{p{1.3cm} | C{0.6cm} C{0.6cm} C{0.6cm} C{0.6cm} C{0.6cm} C{0.6cm} >{\columncolor[gray]{0.8}}C{0.6cm} | C{0.6cm} C{0.6cm} C{0.6cm} C{0.6cm} }
    \rot{} &
    \rot{BERT} & \rot{RoBERTa} & \rot{SimCSE} & \rot{DiffCSE} & \rot{SEED} & \rot{Condenser} & \rot{SentMAE$^{\dag}$} & \rot{GTR (base)} & \rot{GTR (large)} & \rot{GTR (xlarge)} & \rot{Contriever} \\
    \hline
    TREC-COVID & 0.615 & 0.649 & 0.460 & 0.492 & 0.627 & 0.750 & \underline{\textbf{0.772}} & 0.539 & 0.557 & 0.584 & 0.596 \\
    BioASQ & 0.253 & 0.279 & 0.263 & 0.258 & 0.308 & 0.322 & \underline{\textbf{0.421}} & 0.271 & 0.320 & 0.317 & -- \\
    NFCorpus & 0.260 & 0.243 & 0.260 & 0.259 & 0.278 & 0.277 & \underline{0.308} & 0.308 & 0.329& \textbf{0.343} & 0.328 \\
    NQ & 0.467 & 0.413 & 0.435 & 0.412 & 0.446 & 0.486 & \underline{0.518}  & 0.495 & 0.547 & \textbf{0.559} & 0.498 \\
    HotpotQA & 0.488 & 0.448 & 0.502 & 0.499 & 0.541 & 0.538 & \underline{0.635} & 0.535 & 0.579 & 0.591 & \textbf{0.638} \\
    FiQA-2018 & 0.252 & 0.291 & 0.250 & 0.229 & 0.259 & 0.259 & \underline{0.316} & 0.349 & 0.424 & \textbf{0.444} & 0.329  \\
    Signal-1M(RT) & 0.204 & 0.229 & 0.262 & 0.260 & 0.256 & 0.261 & \underline{0.265} & 0.261 & 0.265 & {0.268} & -- \\
    TREC-NEWS & 0.362 & 0.385 & 0.356 & 0.363 & 0.358 & 0.376 & \underline{\textbf{0.428}} & 0.337 & 0.343 & 0.350 & -- \\
    Robust04 & 0.351 & 0.384 & 0.330 & 0.343 & 0.365 & 0.349 & \underline{0.447} & 0.437 & 0.470 & \textbf{0.479} & -- \\
    ArguAna & 0.265 & 0.395 & 0.413 & 0.468 & 0.389 & 0.298 & \underline{0.433} & 0.511 & 0.525 & \textbf{0.531} & 0.446 \\
    Touche-2020 & 0.259 & 0.299 & 0.159 & 0.168 & 0.225 & \underline{\textbf{0.248}} & {0.237} & 0.205 & 0.219 & 0.230 & 0.230 \\
    CQADupStack & 0.282 & 0.278 & 0.290 & 0.305 & 0.290 & \underline{0.347} & 0.317 & 0.357 & 0.384 & \textbf{0.388} & 0.345 \\
    Quora & 0.787 & 0.509 & 0.844 & 0.850 & {0.852} & \underline{0.853} & 0.847 & 0.881 & \textbf{0.890} & \textbf{0.890} & 0.865 \\
    DBPedia & 0.314 & 0.275 & 0.314 & 0.303 & 0.330 & 0.339 & \underline{0.390} & 0.347 & 0.391 & 0.396 & \textbf{0.413} \\
    SCIDOCS & 0.113 & 0.111 & 0.124 & 0.125 & 0.124 & 0.133 & \underline{0.150} & 0.149 & 0.158 & 0.159 & \textbf{0.165} \\
    FEVER & 0.682 & 0.683 & 0.623 & 0.641 & 0.641 & 0.691 & \underline{\textbf{0.774}} & 0.660 & 0.712 & 0.717 & 0.758 \\
    C-FEVER & 0.187 & 0.222 & 0.211 & 0.200 & 0.176 & 0.211 & \underline{0.232} & 0.241 & 0.262 & \textbf{0.270} & 0.237 \\
    SciFact & 0.533 & 0.539 & 0.554 & 0.523 & 0.575 & 0.593 & \underline{0.653} & 0.600 & 0.639 & 0.635 & \textbf{0.677} \\
    \hline
    AVERAGE & 0.371 & 0.368 & 0.369 & 0.372 & 0.391 & 0.407 & \underline{0.452} & 0.416 & 0.445 & \textbf{0.453} & -- \\ 
    \hline
    \end{tabular}
    \caption{Zero-shot dense retrieval performances on BEIR benchmark (measured by NDCG@10). The methods on the left half are based on a BERT-base scale encoding network and pre-trained on Wikipedia and BookCorpus; the methods on the right half are based on enlarged model architectures and substantially increased pre-training data. The highest performances are marked in \textbf{bold}; the highest performances on the left half are \underline{underlined}.}  
    \label{tab:1}
\end{table*}

\subsection{Two-stage Pre-Training}

A two-stage pre-training workflow is conducted as Figure \ref{fig:1}. In the first stage, the model is trained by RetroMAE on generic corpus. We use English Wikipedia and BookCorpus, which are also the pre-training data used by BERT pre-training. We use the following hyper-parameter settings for the first stage pre-training: the encoder is a BERT-base scale transformer and the decoder is a single-layer transformer; the masking ratio is 0.3 for encoder and 0.5 for decoder. Following the practice in BEIR benchmark \citep{thakur2021beir}, the base model from the first stage is fine-tuned with MS MARCO queries for zero-shot retrieval in other domains.  

In the second stage, the base model continues to be trained to deal with more specific downstream tasks. We consider two typical scenarios in this place: in-domain retrieval and sentence embedding for tasks related to semantic matching. The model is trained on MS MARCO corpus \citep{nguyen2016ms} and Natural Questions \citep{kwiatkowski2019natural} corpus respectively, so as to support the in-domain retrieval tasks on the corresponding datasets. Besides, the model is trained with sentence pairs from NLI datasets \citep{bowman2015large,williams2017broad} such that the generated sentence embeddings may serve scenarios like semantic matching. For the second stage, we make use of two different pre-training strategies: on one hand, RetroMAE may continue to be conducted on the domain-specific corpus; on the other hand, the contrastive learning may also be leveraged. The contrastive learning loss $\mathcal{L}_{ctr}$ is formulated by the following equation: 
\begin{equation}
    \mathcal{L}_{ctr} = - \sum\nolimits_{\mathbf{h}_i^+} \log \frac{e^{sim(\mathbf{h}_i,\mathbf{h}_i^+)/\tau}}{\sum_{\mathbf{h}_i^-} e^{sim(sim(\mathbf{h}_i,\mathbf{h}_i^-)/\tau}},
\end{equation}
where $\mathbf{h}_i^+$, $\mathbf{h}_i^-$ are the positive and negative samples to $\mathbf{h}_i$, and $\tau$ is the temperature parameter. While pre-training for in-domain retrieval, we leverage data augmentation to generate the positive samples: following the approach in coCondenser \citep{gao2021unsupervised}, two sentences sampled from the same article are used as $\mathbf{h}_i$ and $\mathbf{h}_i^+$. When targeting on the generation of sentence embeddings, we directly leverage the sentence pairs from NLI datasets as $\mathbf{h}_i$ and $\mathbf{h}_i^+$. For both scenarios, we use in-batch negative samples; that's to say, one training instance (e.g., $\mathbf{h}_i$) will use others' positive samples within the same training batch as its negative samples ($\{\mathbf{h}^+_j\}_{i\neq{j}}$). 

We empirically examine the following combinations of RetroMAE and contrastive learning. For in-domain retrieval, we consider using RetroMAE alone as Eq. \ref{eq:9} (1), and the summation of RetroMAE and contrastive learning as Eq. \ref{eq:9} (2):
\begin{equation}\label{eq:9}
    (1) ~ \underbrace{\mathcal{L}_{enc} + \mathcal{L}_{dec}}_{\text{RetroMAE}}, ~ ~
    (2) ~
    \underbrace{(\mathcal{L}_{enc} + \mathcal{L}_{dec})}_{\text{RetroMAE}} + \mathcal{L}_{ctr}.
\end{equation}
The first option is the simple continuation of RetroMAE on domain-specific corpus. The second option follows the same spirit as coCondenser, except that the pre-training task is switched from Condenser to RetroMAE. While used for the generation of sentence embeddings, the contrastive learning will always be enabled. 

\section{Experiments}
We conduct three sets of experimental studies to evaluate the pre-trained models' effectiveness. (1) For the base model pre-trained on the generic corpus, we evaluate its zero-shot retrieval quality with BEIR benchmark \citep{thakur2021beir}. (2) For the continuingly pre-trained model on domain-specific data, we evaluate the in-domain retrieval quality on MS MARCO \citep{nguyen2016ms} and Natural Questions \citep{kwiatkowski2019natural}. (3) We also evaluate the sentence embeddings' quality based on the semantic textual similarity (STS) \citep{agirre2016semeval,cer2017semeval,marelli2014sick} and transfer tasks in SentEval \citep{conneau2018senteval}. 


\setlength{\tabcolsep}{0.25em} 
\begin{table*}[t]
    \centering
    \tiny
    \begin{tabular}{p{1.2cm}|C{1.0cm}|C{1.1cm}|C{1.0cm}|C{1.0cm}|C{1.0cm}|C{0.9cm}|C{0.9cm}|C{0.9cm}|C{0.9cm}|C{1.0cm} }
    \ChangeRT{1pt} 
    & 
    \multicolumn{5}{c|}{\textbf{MS MARCO}} & \multicolumn{5}{c}{\textbf{Natural Questions}} \\
    \cmidrule(lr){1-1}
    \cmidrule(lr){2-6}
    \cmidrule(lr){7-11}
    \textbf{Methods} & 
    \textbf{MRR@10} & \textbf{MRR@100} & \textbf{R@10} & \textbf{R@100} & \textbf{R@1000} & \textbf{R@10} & \textbf{R@20} & \textbf{R@30} & \textbf{R@50} & \textbf{R@100} \\
    \hline
    BERT & 0.346 & 0.357 & 0.622 & 0.873 & 0.964 & 0.788 & 0.823 & 0.846 & 0.860 & 0.878 \\
    RoBERTa & 0.343 & 0.354 & 0.613 & 0.871 & 0.964 & 0.763 & 0.805 & 0.828 & 0.845 & 0.870\\
    DeBERTa & 0.340 & 0.351 & 0.602 & 0.872 & 0.967 & 0.765 & 0.810 & 0.829 & 0.848 & 0.870 \\
    SimCSE & 0.352 & 0.362 & 0.628 & 0.885 & 0.974 & 0.774 & 0.819 & 0.842 & 0.863 & 0.886 \\ 
    LaPraDoR & 0.346 & 0.356 & 0.613 & 0.876 & 0.964 & 0.780 & 0.825 & 0.842 & 0.859 & 0.877 \\
    DiffCSE & 0.346 & 0.357 & 0.622 & 0.875 & 0.965 & 0.785 & 0.825 & 0.841 & 0.862 & 0.878 \\
    SEED & 0.354 & 0.365 & 0.626 & 0.881 & 0.969 & 0.780 & 0.826 & 0.845 & 0.868 & 0.887 \\
    Condenser & 0.364 & 0.374 & 0.639 & 0.891 & 0.972 & 0.790 & 0.833 & 0.852 & 0.867 & 0.883 \\
    SentMAE$^\dag$ & {0.378} & {0.389} & {0.666} & {0.909} & {0.980} & {0.804} & {0.844} & \textbf{0.863} & \textbf{0.878} & {0.894} \\
    \hhline{=|=|=|=|=|=|=|=|=|=|=} 
    Co-Conderser & 0.382 & -- & -- & -- & 0.984 & -- & 0.843 & -- & -- & 0.890 \\ 
    SentMAE$^{\ddag}$ & \textbf{0.393} & \textbf{0.403} & \textbf{0.675} & \textbf{0.918} & \textbf{0.985} & \textbf{0.805} & \textbf{0845} & \textbf{0.863} & \textbf{0.878} & \textbf{0.895} \\
    SentMAE$^\ddag_{co}$ & 0.389 & 0.398 & 0.674 & 0.915 & \textbf{0.985} & -- & -- & -- & -- & -- \\
    \ChangeRT{1pt}
    \end{tabular}
    \caption{In-domain retrieval based on ANCE fine-tuning. The models in the upper half are pre-trained with Wikipedia and BookCorpus; the models in the lower half are further pre-trained with domain-specific data. SentMAE$^{\ddag}$ purely relies on RetroMAE during the second-stage pre-training; while SentMAE$^{\ddag}_{co}$ jointly employs RetroMAE and contrastive learning.} 
    \label{tab:2}
\end{table*}  

\subsection{Zero-shot Retrieval}

The zero-shot retrieval performances on BEIR benchmark are shown with Table \ref{tab:1}. The methods on the left half of the table follow similar settings: a BERT-base scale transformer (110M model parameters) is used as the encoding network, and the pre-training is performed mainly based on Wikipedia and BookCorpus (except RoBERTa). Among these methods: BERT \cite{Devlin2019BERT} and RoBERTa \citep{Liu2019Roberta} are the pre-trained models for generic NLU tasks; SimCSE \citep{gao2021simcse} and DiffCSE \citep{chuang2022diffcse} are two popular methods for pre-trained sentence embedding; SEED \citep{lu2021less} and Condenser \citep{gao2021condenser} are competitive pre-trained models for dense retrieval. SentMAE$^\dag$ stands for the reproduction of RetroMAE. 

We can observe that SentMAE$^\dag$ achieves overwhelming advantages over the above pre-trained models under the same settings. There's a huge improvement: 0.452 over 0.407 (the best of baselines) in total average, and it leads to the best retrieval quality on the majority of the datasets. Such an observation verifies RetroMAE's effectiveness for zero-shot retrieval. Besides, RoBERTa's performance is not better than BERT; and pre-trained models for sentence embeddings, like SimCSE and DiffCSE, do not contribute to the zero-shot retrieval.  

A number of the competitive pre-trained models on BEIR benchmark are listed on the right side of Table \ref{tab:1} for further comparison. These models take advantages of enlarged encoding networks and substantially increased pre-training data to enhance their cross-domain retrieval capability. 

Specifically, GTR \citep{ni2021large} is pre-trained based on contrastive learning on a massive dataset of 2 billion community question-answer pairs, and it employs transformers of different scales: GTR-base (110M), GTR-large (335M), GTR-XL (1.24B), GTR-XXL (4.8B), as the encoding networks. Although working with limited data and model size, RetroMAE still outperforms GTR-base (both with 110M model parameters) by a big margin; and its overall performance is on-par with GTR-xlarge, whose model size is more than 11$\times$ larger.  

Another strong zero-shot retriever is Contriever \citep{izacard2021towards}, whose overall performance (0.466 in total average) is slightly lower than SentMAE$^\dag$ (averagely 0.470 on the datasets used by Contriever). It is pre-trained via self-contrastive learning, which takes place on Wikipedia and CC-Net \citep{wenzek2020ccnet}. The second dataset consists of about 700 millions high quality documents, which is more than 100$\times$ larger than the data used for SentMAE$^\dag$'s pre-training. It also requires sophisticated data augmentation and a considerable volume of negative samples (as many as 131,072) leveraging MoCo \citep{he2020momentum}. By comparison, SentMAE$^\dag$ is simple and cost-friendly, as it uses much less pre-training data and requires no negative samples.

\setlength{\tabcolsep}{0.1em} 
\begin{table*}[t]
    \centering
    \tiny
    \begin{tabular}{p{1.5cm}|C{1.2cm}C{1.2cm}C{1.3cm}|p{1.5cm}|C{1.2cm}C{1.2cm}C{1.2cm} } 
    \ChangeRT{1pt} 
    \textbf{Dataset} & \textbf{SentMAE$^{\dag}$} & \textbf{SentMAE$^{\ddag}$} & \textbf{SentMAE$^{\S}$} &\textbf{Dataset} & \textbf{SentMAE$^{\dag}$} & \textbf{SentMAE$^{\ddag}$} & \textbf{SentMAE$^{\S}$} \\
    \hline
    TREC-COVID & \textbf{0.772} & 0.652 & 0.673 & Signal-1M(RT) & {0.265} & 0.258 & \textbf{0.279} \\
    Quora & 0.847 & \textbf{0.853} & 0.846 & BioASQ & \textbf{0.421} & 0.367 & 0.373 \\ 
    TREC-NEWS & \textbf{0.428} & 0.348 & 0.365 & DBPedia & {0.390} & 0.355 & \textbf{0.405} \\
    NFCorpus & 0.308 & \textbf{0.327} & 0.307 & Robust04 & \textbf{0.447} & 0.392 & 0.411 \\ 
    SCIDOCS & \textbf{0.150} & 0.133 & 0.146 & NQ & {0.518} & 0.483 & \textbf{0.529} \\ 
    ArguAna & \textbf{0.433} & 0.247 & {0.315} & FEVER & \textbf{0.774} & 0.685 & 0.722 \\ 
    HotpotQA & \textbf{0.635} & 0.590 & 0.634 & Touche-2020 & \textbf{0.347} & 0.258 & 0.209 \\ 
    C-FEVER & \textbf{0.232} & 0.171 & 0.209 & FiQA-2018 & \textbf{0.316} & 0.289 & 0.306 \\ 
    CQADupStack & {0.317} & 0.310 & \textbf{0.322} & SciFact & \textbf{0.653} & 0.634 & \textbf{0.653} \\
    \hline
    \textbf{Average} & \textbf{0.452} & 0.408 & 0.428 & \multicolumn{4}{c}{} \\
    \ChangeRT{1pt} 
    \end{tabular}
    \caption{Comparisons of zero-shot retrieval between SentMAE$^{\dag}$: the base model pre-trained on generic corpus with RetroMAE, {SentMAE$^{\ddag}$}: the continuingly pre-trained model on MS MARCO corpus, and SentMAE$^{\S}$: the continuingly pre-trained model on Natural Questions corpus (Wikipedia).} 
    \label{tab:3}
\end{table*}

\subsection{In-domain Retrieval}

The evaluation of in-domain retrieval is shown as Table \ref{tab:2}. The upper part of the table demonstrates the performances from models pre-trained on generic corpus; the lower part of the table indicates the performances from the models pre-trained on domain-specific data. We can derive the following interesting points from the demonstrated results. SentMAE$^{\ddag}$ and SentMAE$^{\ddag}_{co}$ indicate stand for RetroMAE only and RetroMAE plus contrastive learning in the second stage pre-training. 

First, the continued pre-training on domain-specific data is important to the in-domain retrieval quality. Although the first-stage pre-training already equips the model with a strong capability on dense retrieval, as SentMAE$^{\dag}$ outperforms all baselines in the upper half, the continued in-domain pre-training further brings substantially benefits. By keeping on the RetroMAE pre-training on MS MARCO corpus, SentMAE$^{\ddag}$ achieves significant improvements over the base model SentMAE$^{\dag}$, which is pre-trained on the generic data. By comparison, the improvement is relatively marginal on Natural Questions. This is probably because the NQ's corpus, i.e., Wikipedia, is already included as a part of the generic pre-training data. The corresponding in-domain retrieval quality from SentMAE$^{\ddag}$ is indeed competitive: the performances are almost the same as the ones based on sophisticated knowledge distillation \citep{ren2021rocketqav2} and adversarial learning \citep{zhang2021adversarial}, and might be the strongest performances with simple ANCE fine-tuning on both benchmarks. 

Second, we explore which learning strategy, RetroMAE or self-contrastive learning, leads to the improvement of retrieval quality. According to our current experiments, the self-contrastive does not contribute to the in-domain retrieval quality, as performances from SentMAE$^{\ddag}_{co}$ are no better than SentMAE$^{\ddag}$. To further confirm this point, we ablate the contrastive learning part from coCondenser, and only preserve the in-domain pre-training. The ablation leads to MRR@10 of 0.383 and Recall@1000 of 0.983. Both results are no worse than the original results from coCondenser, which are consistent with the foregoing observations. 

We conjecture that it is the quality of relevance label that restricts the performance of contrastive learning. Particularly, for supervised contrastive learning, such as the one based on the query-passage pairs from MS MARCO or the one with sentence pairs from NLI, the positive labels are reliable and meaningful. Therefore, it will not only benefit the in-domain performance, but also contribute to the transferability towards other domains. However, for self-contrastive learning, the labeled relevance comes from data augmentation, which is likely to be either over simplified to be discriminated, or introduce potential false positives. 

Third, the pre-training on domain-specific data can be harmful to the generic retrieval quality in other domains. We explore the impacts introduced from SentMAE$^{\ddag}$ and SentMAE$^{\S}$, which are continued pre-trained models on MS MARCO and NQ corpus, respectively. As shown in Table \ref{tab:3}, there are notable decrements of zero-shot retrieval quality for the two in-domain pre-trained models; especially SentMAE$^{\ddag}$, which performs continued pre-training on MS MARCO corpus. 

It's worth noting that the pre-trained models are fine-tuned with the query-passage relevance from MS MARCO before they are tested on BEIR benchmark. In other words, MS MARCO is used as the source domain and other datasets in BEIR are treated as the target domains. Intuitively, people may expect that a model will probably work better on target domains if its performance can be substantially improved on the source domain. However, the comparison between SentMAE$^{\dag}$ and SentMAE$^{\ddag}$ is a clear contradiction to such an expectation. It suggests that the retrieval model's transferability might be more of the result 
from pre-training, but less of the fine-tuned performance on the source domain.

\begin{table*}[t]
    \centering
    \tiny
    \begin{tabular}{p{1.6cm}|C{1.2cm}|C{1.2cm}|C{1.2cm}|C{1.2cm}|C{1.2cm}|C{1.2cm}|C{1.2cm}|C{1.2cm} } 
    \ChangeRT{1pt} 
    & 
    \multicolumn{8}{c}{\textbf{Semantic textual similarity}} \\
    \cmidrule(lr){1-1}
    \cmidrule(lr){2-9}
    \textbf{Methods} & 
    \textbf{STS12} & \textbf{STS13} & \textbf{STS14} & \textbf{STS15} & \textbf{STS16} & \textbf{STS-B} & \textbf{STS-R} & \textbf{Average} \\
    \hline
    SimCSE-BERT & 75.30 & 84.67 & 80.19 & 85.40 & 80.82 & 84.25 & \textbf{80.39} & 81.57 \\
    SentMAE$^{\mathparagraph}$ & \textbf{76.83} & \textbf{86.20} & \textbf{81.42} & \textbf{86.59} & \textbf{82.25} & \textbf{85.08} & 79.80 & \textbf{82.60} \\ 
    \hline
    & 
    \multicolumn{8}{c}{\textbf{SentEval Transfer Tasks}} \\
    \cmidrule(lr){1-1}
    \cmidrule(lr){2-9}
    \textbf{Methods} & 
    \textbf{MR} & \textbf{CR} & \textbf{SUBJ} & \textbf{MPQA} & \textbf{SST} & \textbf{TREC} & \textbf{MRPC} & \textbf{Average} \\
    \hline
    SimCSE-BERT & \textbf{82.69} & \textbf{89.25} & \textbf{94.81} & 89.59 & \textbf{87.31} & 88.40 & 73.51 & 86.51 \\
    SentMAE$^{\mathparagraph}$ & 82.29 & 88.88 & 94.64 & \textbf{90.29} & 86.88 & \textbf{92.00} & \textbf{75.07} & \textbf{87.15} \\ 
    \ChangeRT{1pt}
    \end{tabular}
    \caption{Evaluation of sentence embedding quality. The upper table indicates the performances on semantic textual similarity; the lower table indicates performances on the transfer tasks in SentEval.} 
    \label{tab:4}
\end{table*}

\subsection{Sentence Embeddings}

The evaluation of sentence embedding quality is shown with Table \ref{tab:4}, where SentMAE$^{\mathparagraph}$ is compared with SimCSE-BERT (base) \citep{gao2021simcse}. The two methods rely on the same contrastive learning process, i.e., the supervised contrastive learning with sentence pairs from NLI datasets. However, SentMAE$^{\mathparagraph}$ leverages RetroMAE for pre-training, whose impact will be reflected from the comparison. 

It can be observed from the experiment results that the pre-training from RetroMAE indeed contributes to the sentence embedding quality. It achieves substantial empirical gains on most of the STS tasks, where the overall performance can be improved from 81.57\% to 82.60\%. The average performance of transfer tasks may also be improved from 86.51\% to 87.15\%, despite that it is not as significant as the gain on the STS tasks. It is probably because the improvement of sentence representation may not directly benefit the transfer tasks, as discussed in \citep{gao2021simcse}. 

Notice that the previous works seldom evaluate a pre-trained model simultaneously on its capability as a dense retriever and the quality of its sentence embeddings. This is probably because the two tasks are about different intrinsic properties of the pre-trained models. In other words, a pre-trained model which is effective for one type of task may become ineffective for another task. For example, SimCSE brings no benefit to zero-shot and in-domain retrieval according to our observations in Table \ref{tab:1} and \ref{tab:2}, despite its effectiveness in STS tasks. However, given what we have 
found in this work, both tasks may benefit from the pre-training of RetroMAE. 

\subsection{Summarization}
The major experimental findings in this work can be summarized into the following points. 


$\bullet$ The pre-training of RetroMAE on generic corpus contributes to all of the tasks in experiments, including zero-shot retrieval, in-domain retrieval, sentence embeddings for semantic textual similarity and transfer tasks in SentEval. 

$\bullet$ The continued pre-training on domain-specific data substantially benefits the in-domain retrieval performances. However, it can be harmful to the model's transferability, as the zero-shot retrieval quality decreases for the out-domain data.  

$\bullet$ The contrastive learning's impact is conditional in our experiments. It helps to generate high-quality sentence embeddings for STS and transfer tasks in SentEval; however, it turns out to be relatively limited by its sample efficiency for zero-shot retrieval, and it does not bring extra benefit to in-domain retrieval at the presence of RetroMAE. 

\section{Conclusion}
In this paper, a unified framework is presented for the pre-training of sentence representation. On top of the consecutive MAE style pre-training on generic and domain-specific data, and with proper collaboration with contrastive learning, the generated model may support a wide variety of sentence representation tasks, including zero-shot retrieval, in-domain retrieval, and sentence embeddings. The experiment studies demonstrate that the proposed framework helps to achieve strong performances on benchmarks, like BEIR, MS MARCO, Natural Questions, STS and transfer tasks in SentEval. For future works, we'll make extensions by scaling up the size of the encoding network and having it pre-trained on more unsupervised data, which will push to the empirical limit of the current method. We'll also explore its effectiveness in other languages other than English, and its impact to more applications beyond dense retrieval and NLI. 

\clearpage

\bibliographystyle{acl_natbib}
\bibliography{main}

\end{document}